\documentclass{article}

\PassOptionsToPackage{numbers, compress}{natbib}
\usepackage[dblblindworkshop, final]{neurips_2025}

\usepackage[utf8]{inputenc} % allow utf-8 input
\usepackage[T1]{fontenc}    % use 8-bit T1 fonts
\usepackage{hyperref}       % hyperlinks
\usepackage{url}            % simple URL typesetting
\usepackage{booktabs}       % professional-quality tables
\usepackage{amsfonts}       % blackboard math symbols
\usepackage{nicefrac}       % compact symbols for 1/2, etc.
\usepackage{microtype}      % microtypography
\usepackage{xcolor}         % colors
\usepackage{graphicx}
\usepackage{siunitx}
\usepackage{placeins}
\usepackage{float}

\title{BeetleFlow: An Integrative Deep Learning Pipeline for Beetle Image Processing}

\author{
\textbf{Fangxun~Liu}$^{1*}$, \textbf{S~M~Rayeed}$^{2*}$, \\
Samuel~Stevens$^{1}$, Alyson~East$^{3}$, Cheng~Hsuan~Chiang$^{1}$, Colin~Lee$^{1}$, Daniel~Yi$^{1}$, Junke~Yang$^{1}$, \\
Tejas~Naik$^{1}$, Ziyi~Wang$^{1}$, Connor~Kilrain$^{1}$, Elijah~H.~Buckwalter$^{1}$, Jiacheng~Hou$^{1}$, \\
Saul~Ibaven~Bueno$^{1}$, Shuheng~Wang$^{1}$, Xinyue~Ma$^{1}$, Yifan~Liu$^{1}$, Zhiyuan~Tao$^{1}$, Ziheng~Zhang$^{1}$, \\
Eric~Sokol$^{4}$, Michael~Belitz$^{5}$, Sydne~Record$^{3}$, Charles~V.~Stewart$^{2}$, Wei-Lun~Chao$^{1}$ \\[.5em]
$^{1}$The Ohio State University \quad
$^{2}$Rensselaer Polytechnic Institute \quad
$^{3}$The University of Maine \\
$^{4}$National Ecological Observatory Network (NEON), Battelle\quad
$^{5}$Michigan State University \\[.5em]
\small{*Corresponding authors: \texttt{liu.12122@osu.edu}, \texttt{rayees@rpi.edu}}
}
\begin{document}

\maketitle

\begin{abstract}
  In entomology and ecology research, biologists often need to collect a large number of insects, among which beetles are the most common species. A common practice for biologists to organize beetles is to place them on trays and take a picture of each tray. Given the images of thousands of such trays, it is important to have an automated pipeline to process the large-scale data for further research. Therefore, we develop a 3-stage pipeline to detect all the beetles on each tray, sort and crop the image of each beetle, and do morphological segmentation on the cropped beetles. For detection, we design an iterative process utilizing a transformer-based open-vocabulary object detector and a vision-language model. For segmentation, we manually labeled 670 beetle images and fine-tuned two variants of a transformer-based segmentation model to achieve fine-grained segmentation of beetles with relatively high accuracy. The pipeline integrates multiple deep learning methods and is specialized for beetle image processing, which can greatly improve the efficiency to process large-scale beetle data and accelerate biological research.
\end{abstract}
\section{Introduction}
\label{introduction}

In entomology and ecology research, biologists often need to collect a large number of insects to study, among which beetles are one of the most common species. Beetles account for around 25\% of all known species in the world~\cite{Hammond1992}, therefore, they have significant research value in a variety of fields such as evolution and biodiversity, for their species richness and widespread distribution. In practice, biologists often use pins to mount beetles collected at the same site and time on a tray. A tray contains from several to over 60 beetle specimens. After collection, biologists can have up to thousands of trays and take a picture of each tray to digitize. This procedure brings in a large amount of beetle data organized by trays, but how to process them for further study becomes a new question.

Given the importance of beetle study and the challenges, we develop a 3-stage deep learning pipeline to process large-scale beetle images. In the first stage, we utilize an open-vocabulary detector, Grounding DINO~\cite{liu2024grounding}, for beetle detection and a vision-language model, LLaVA-NeXT~\cite{liu2024llavanext}, for final verification. This approach achieves a high accuracy of 97.81\%. In the second stage, we crop each detected beetle from the tray image and save it as a single image, with optional sorting and metadata matching. In the third stage, we leverage a transformer-based model, Mask2Former~\cite{cheng2021mask2former}, to segment each beetle into 5 or 9 morphological parts. The model achieves a mean Intersection over Union (mIOU) of 85.11\% for 5 class segmentation and 77.38\% for 9 class segmentation.

Multiple downstream research on beetles can be conducted based on the high-throughput data processed by our pipeline. Moreover, the pipeline has the potential to generalize to more biological data processing cases, as we have observed a similar detect-and-segment pattern in other biological pipelines, such as QuPath~\cite{bankhead2017qupath} for cells and PlantCV~\cite{gehan2017plantcv} for plants. This pattern is applicable to a variety of biological data processing workflows, and our work on beetles sets an example for insects, which are one of the most numerous groups of organisms in the world.
\section{Related Work}
\label{related-work}

While traditional object detectors like the R-CNN family~\cite{girshick2014rich} and YOLO~\cite{redmon2016you} are limited by predefined datasets, open-vocabulary detection methods~\cite{li2022grounded, gu2021open} now leverage language prompts to detect arbitrary objects. Similarly, Vision-Language Models (VLMs)~\cite{radford2021learning, li2023blip} extend the reasoning of large language models~\cite{achiam2023gpt, grattafiori2024llama, team2024gemini} to the multimodal domain for joint text-image reasoning. In parallel, the state-of-the-art in semantic segmentation has shifted from Convolutional Neural Networks (CNNs) like U-Net~\cite{ronneberger2015u} to transformer-based models~\cite{vaswani2017attention, dosovitskiy2020image, liu2021swin}. These models leverage self-attention for global context, which is critical for distinguishing morphologically similar parts in biological imaging. These advancements are relevant to beetle studies, where machine learning, supported by large-scale beetle sampling programs like NEON~\cite{hoekman2017design}, is already applied for identification and classification. Existing works range from traditional algorithms on extracted features~\cite{blair2020robust} to deep learning methods, such as CNNs for identification~\cite{wu2019deep} and deep vision models for fine-grained taxonomic classification~\cite{rayeed2025fine, rayeed2025beetleverse}.
\section{Beetle Image Processing Pipeline}
\label{pipeline}

\begin{figure}[t]
  \centering
  \includegraphics[width=0.9\columnwidth]{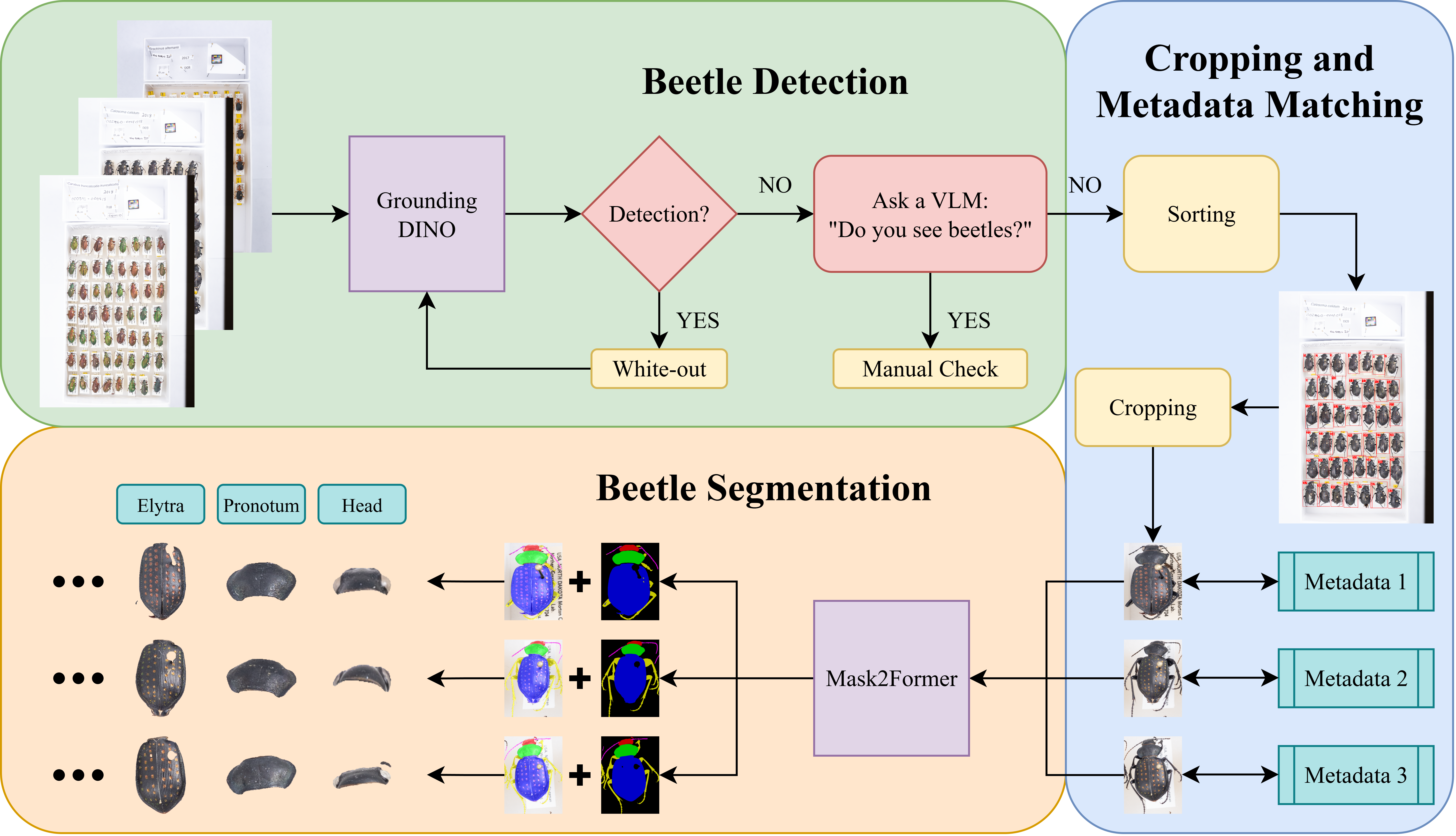}
  \caption{An overview of the 3-stage pipeline: individual detection, cropping/metadata matching, and body-part segmentation stages.}
  \label{fig:pipeline}
\end{figure}

The input to our pipeline is a series of images of trays containing multiple beetles. Each tray image undergoes a 3-stage processing (Figure~\ref{fig:pipeline}). The code and data are available at \url{https://github.com/Imageomics/BeetleFlow.git}.

\subsection{Iterative Beetle Detection}

The first stage is iterative beetle detection. In one iteration, the tray image and the text prompt ``a beetle'' are sent as input to Grounding DINO, which then outputs the bounding box coordinates for the detected beetles. Next, white masks are placed over all detected beetles based on the bounding boxes, leaving the undetected ones in the tray image. The resulting modified image then proceeds to the next iteration for another round of detection and masking. When no detection is reported, the iteration stops. The final modified image is then sent to LLaVA-NeXT with the text prompt ``Do you see beetles in this image?''. We constrain the model to output ``YES'' or ``NO'' as the final word for automated check. If it answers ``YES'', a message is sent to the user to check manually. If it answers ``NO'', the detection is successful and the bounding box coordinates are sent to the next stage.

\subsection{Beetle Image Cropping}

The second stage takes the bounding box coordinates as input and outputs individual cropped beetle images, with optional functions of sorting and metadata matching. Given the bounding box coordinates, the pipeline crops the beetles out of the original tray image and saves them as individual images. A specific order can be applied when saving the beetle images. In practice, the beetles on a tray are arranged in regular rows and columns. If digitized metadata for the beetles are available, they are typically provided in a left-to-right, top-to-bottom order by the biologists. The pipeline sorts the beetle images according to the top-left bounding box coordinate and associates the metadata with each beetle. The metadata matched to the beetles are saved in a CSV file for each tray.

\subsection{Fine-grained Beetle Segmentation}

The third stage takes individual beetle images as input and outputs the morphological segmentation results for each beetle. For this task, we fine-tune two variants of Mask2Former, based on granularity: a 5-class model (segmenting head, pronotum, elytra, legs, and antennas) for basic morphological analysis, and a 9-class model (additionally eyes, mouthparts, tail, and pin). Both models inherently separate the beetle from the background. For each input image, the model generates a colorized mask image and an image overlaid with the masks. The overlaid image is provided for user verification. The mask image can be utilized for morphological part cropping and defective specimen detection.
\section{Experiments}
\label{experiments}

\subsection{Experimental Setup}

\subsubsection{Beetle Detection}

\paragraph{Datasets.}
We apply the detection pipeline on the dataset collected by the National Ecological Observatory Network (NEON)~\cite{NEON-pinned-beetles, NEON-field-metadata} from ecological sites across the U.S., along with associated metadata. The dataset contains 1,506 images of trays containing pinned carabid specimens. The beetles were imaged following the optimized image capture guidelines for biodiversity specimens~\cite{east2025optimizing}.

\paragraph{Evaluation metrics.}
Each tray in the NEON dataset is associated with respective metadata, including the ground-truth number of beetles on the tray. After running the detection process, the number of detected beetles is compared to the ground-truth number. The exact-match accuracy over the 1,506 tray images is then calculated to quantify the model's performance.

\paragraph{Implementation Details.}

We utilized the Grounding DINO model with pre-trained weights from the \texttt{IDEA-Research/grounding-dino-base} checkpoint. A fixed text prompt, ``a beetle.'', was used to guide the model in locating specimens. For post-processing, we set the box confidence threshold to 0.3 and the text relevance threshold to 0.2. Only bounding boxes with scores exceeding both respective thresholds are retained. We utilized the LLaVA-NeXT model with pre-trained weights from the \texttt{llava-hf/llava-v1.6-mistral-7b-hf} checkpoint for the final verification.

\subsubsection{Beetle Segmentation}

\paragraph{Datasets.}
The unlabeled individual beetle images are derived from our pipeline by processing the tray images. For 5-class labeling, we manually labeled 160 beetles and utilized an additional 180 labeled beetles from a previous work, SST~\cite{feng2025sst, Fluck2018_NEON_Beetle, Ramirez_2018_NEON_Ethanol-preserved_2025}. A total of 340 labeled beetles were then partitioned into a training set of 272 and a test set of 68 images. For 9-class labeling, we manually labeled 330 beetles, dividing them into a training set of 264 and a test set of 66 images.

\paragraph{Evaluation metrics.}
We use mIOU on the test set as the primary metric, averaged across all classes for each image. We also report some per-class IoU scores to facilitate a more granular analysis.

\paragraph{Implementation Details.}
We fine-tuned two Mask2Former models with a Swin-Large backbone, each initialized with weights from the \texttt{facebook/mask2former-swin-large-ade-semantic} checkpoint, which was pre-trained on the ADE20K dataset for semantic segmentation tasks. All input images were resized to a resolution of 512$\times$512 pixels. The models were trained for 30 epochs with a batch size of 10, using the AdamW optimizer and an initial learning rate of 1e-4.

\subsection{Results}

\paragraph{Beetle Detection.}
We applied the detection process to 1,506 tray images. Of these, 1,473 trays had a perfect match between the number of detected beetles and the ground truth, yielding a total accuracy of 97.81\%. Of the 33 failure cases, 32 had a higher detected beetle count than the ground truth, while only one case had a lower count. A majority of these 32 cases were due to fallen beetle heads on the trays, which the model incorrectly detected as separate beetles. The data indicates that our detection process is highly effective at detecting all beetles on a tray with minimal omissions.

\begin{figure}[t]
  \centering
  \includegraphics[width=0.9\columnwidth]{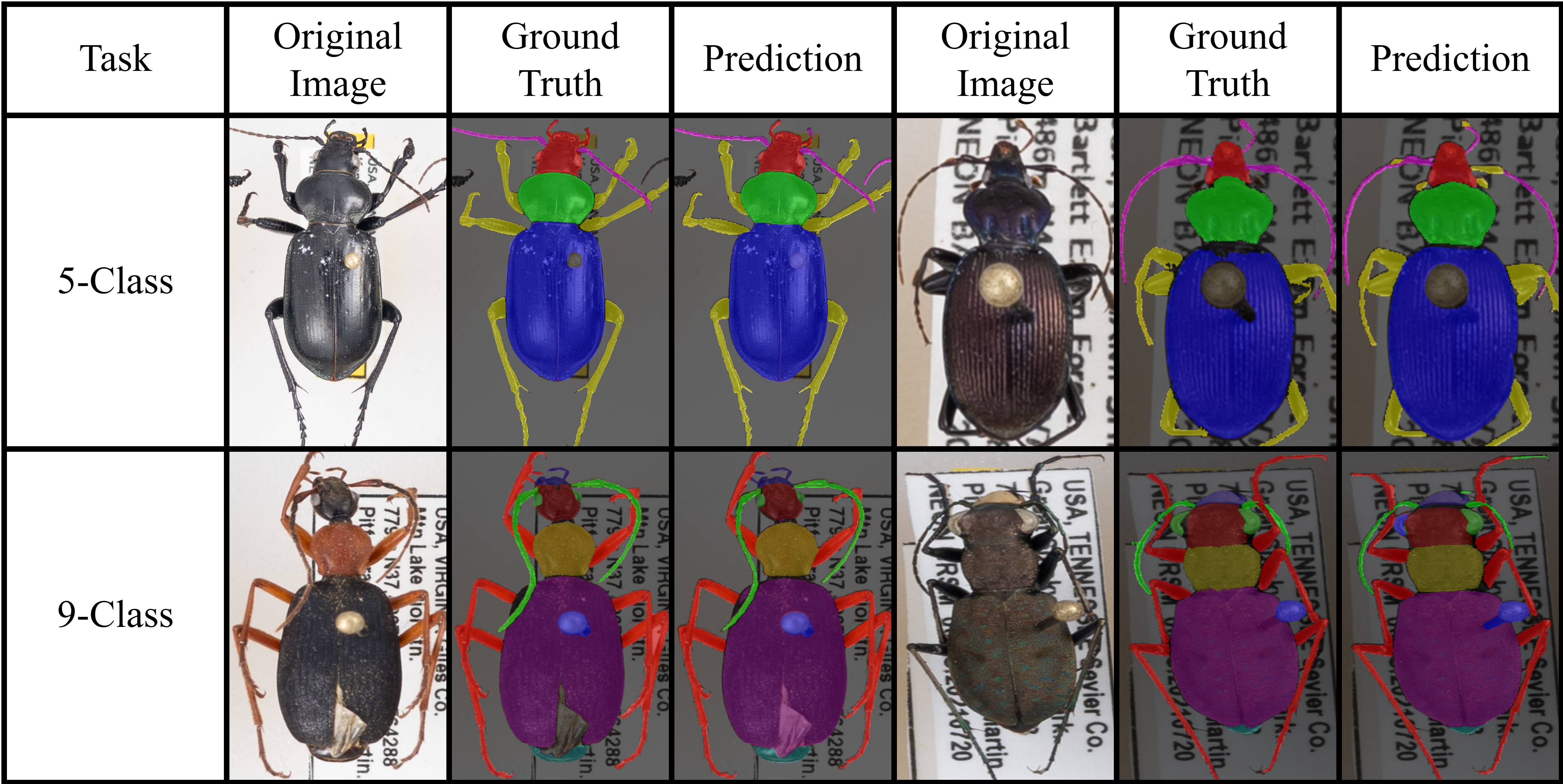}
  \caption{Qualitative segmentation results. Top row: 5-class segmentation (head, pronotum, elytra, legs, antennas). Bottom row: 9-class segmentation (adding eyes, mouthparts, tail, pin). Columns show original, ground truth, and prediction.}
  \label{fig:results}
\end{figure}

\paragraph{Beetle Segmentation.}
We evaluated the performance of two models on their respective test sets. The mIOU is 85.11\% for 5-class segmentation and 77.38\% for 9-class segmentation. For per-class IOU results, the model achieves high scores on large morphological parts like ``pronotum'' (91.85\% and 90.99\%) and ``elytra'' (94.69\% and 93.97\%), while the scores for smaller parts like ``legs'' (79.57\% and 85.39\%) and ``antennas'' (65.93\% and 70.08\%) are lower. This phenomenon is partly attributable to the sensitivity of IoU to object size. For small objects, deviations of a few pixels can lead to a significant drop. Qualitative results (Figure~\ref{fig:results}) show that the segmentation is reasonably good.

\section{Discussion and Future Work}
\label{discussion}

In this work, we develop an automated pipeline to process large-scale beetle images. The two deep learning-based processes, Grounding DINO detection and Mask2Former segmentation, have proven to be highly accurate. In addition, each tray also contains a scale bar and a color table, and we have developed the functionality to detect and crop them. The scale bar can be used to automatically measure the morphological statistics like length and area, and the color table can be used for the color calibration of images. These applications can be added in future work. In summary, our pipeline greatly improves the efficiency of large-scale beetle image processing, yielding useful outputs for downstream research. This pipeline scheme can be further generalized to other organisms beyond beetles, with the potential to improve the data processing workflow in the biology field.

\ack
This work is supported by the National Science Foundation OAC 2118240 Imageomics Institute award and is based in part upon work supported by the National Ecological Observatory Network (NEON), a program sponsored by the U.S. National Science Foundation (NSF) and operated under cooperative agreement by Battelle. The authors are grateful for the generous support of the computational resources from the Ohio Supercomputer Center.

\bibliography{references}

@incollection{Hammond1992,
  author = {Hammond, P. M.},
  title = {Species inventory},
  booktitle = {Global Biodiversity: Status of the Earth's Living Resources. A Report Compiled by the World Conservation Monitoring Centre},
  editor = {Groombridge, Brian},
  publisher = {Chapman and Hall},
  address = {London},
  pages = {17--39},
  year = {1992}
}

@inproceedings{liu2024grounding,
  title={Grounding dino: Marrying dino with grounded pre-training for open-set object detection},
  author={Liu, Shilong and Zeng, Zhaoyang and Ren, Tianhe and Li, Feng and Zhang, Hao and Yang, Jie and Jiang, Qing and Li, Chunyuan and Yang, Jianwei and Su, Hang and others},
  booktitle={European conference on computer vision},
  pages={38--55},
  year={2024},
  organization={Springer}
}

@misc{liu2024llavanext,
    title={LLaVA-NeXT: Improved reasoning, OCR, and world knowledge},
    url={https://llava-vl.github.io/blog/2024-01-30-llava-next/},
    author={Liu, Haotian and Li, Chunyuan and Li, Yuheng and Li, Bo and Zhang, Yuanhan and Shen, Sheng and Lee, Yong Jae},
    month={January},
    year={2024}
}

@inproceedings{cheng2021mask2former,
  title={Masked-attention Mask Transformer for Universal Image Segmentation},
  author={Bowen Cheng and Ishan Misra and Alexander G. Schwing and Alexander Kirillov and Rohit Girdhar},
  booktitle={Proceedings of the IEEE/CVF conference on computer vision and pattern recognition},
  pages={1290--1299},
  year={2022}
}

@article{bankhead2017qupath,
  title={QuPath: Open source software for digital pathology image analysis},
  author={Bankhead, Peter and Loughrey, Maurice B and Fern{\'a}ndez, Jos{\'e} A and Dombrowski, Yvonne and McArt, Darragh G and Dunne, Philip D and McQuaid, Stephen and Gray, Ronan T and Murray, Liam J and Coleman, Helen G and others},
  journal={Scientific reports},
  volume={7},
  number={1},
  pages={1--7},
  year={2017},
  publisher={Nature Publishing Group}
}

@article{gehan2017plantcv,
  title={PlantCV v2: Image analysis software for high-throughput plant phenotyping},
  author={Gehan, Malia A and Fahlgren, Noah and Abbasi, Arash and Berry, Jeffrey C and Callen, Steven T and Chavez, Leonardo and Doust, Andrew N and Feldman, Max J and Gilbert, Kerrigan B and Hodge, John G and others},
  journal={PeerJ},
  volume={5},
  pages={e4088},
  year={2017},
  publisher={PeerJ Inc.}
}

@inproceedings{girshick2014rich,
  title={Rich feature hierarchies for accurate object detection and semantic segmentation},
  author={Girshick, Ross and Donahue, Jeff and Darrell, Trevor and Malik, Jitendra},
  booktitle={Proceedings of the IEEE conference on computer vision and pattern recognition},
  pages={580--587},
  year={2014}
}

@inproceedings{redmon2016you,
  title={You only look once: Unified, real-time object detection},
  author={Redmon, Joseph and Divvala, Santosh and Girshick, Ross and Farhadi, Ali},
  booktitle={Proceedings of the IEEE conference on computer vision and pattern recognition},
  pages={779--788},
  year={2016}
}

@article{achiam2023gpt,
  title={Gpt-4 technical report},
  author={Achiam, Josh and Adler, Steven and Agarwal, Sandhini and Ahmad, Lama and Akkaya, Ilge and Aleman, Florencia Leoni and Almeida, Diogo and Altenschmidt, Janko and Altman, Sam and Anadkat, Shyamal and others},
  journal={arXiv preprint arXiv:2303.08774},
  year={2023}
}

@inproceedings{ronneberger2015u,
  title={U-net: Convolutional networks for biomedical image segmentation},
  author={Ronneberger, Olaf and Fischer, Philipp and Brox, Thomas},
  booktitle={International Conference on Medical image computing and computer-assisted intervention},
  pages={234--241},
  year={2015},
  organization={Springer}
}

@misc{feng2025sst,
      title={Static Segmentation by Tracking: A Frustratingly Label-Efficient Approach to Fine-Grained Segmentation}, 
      author={Zhenyang Feng and Zihe Wang and Saul Ibaven Bueno and Tomasz Frelek and Advikaa Ramesh and Jingyan Bai and Lemeng Wang and Zanming Huang and Jianyang Gu and Jinsu Yoo and Tai-Yu Pan and Arpita Chowdhury and Michelle Ramirez and Elizabeth G. Campolongo and Matthew J. Thompson and Christopher G. Lawrence and Sydne Record and Neil Rosser and Anuj Karpatne and Daniel Rubenstein and Hilmar Lapp and Charles V. Stewart and Tanya Berger-Wolf and Yu Su and Wei-Lun Chao},
      year={2025},
      eprint={2501.06749},
      archivePrefix={arXiv},
      primaryClass={cs.CV},
      url={https://arxiv.org/abs/2501.06749}, 
}

@inproceedings{li2022grounded,
  title={Grounded language-image pre-training},
  author={Li, Liunian Harold and Zhang, Pengchuan and Zhang, Haotian and Yang, Jianwei and Li, Chunyuan and Zhong, Yiwu and Wang, Lijuan and Yuan, Lu and Zhang, Lei and Hwang, Jenq-Neng and others},
  booktitle={Proceedings of the IEEE/CVF conference on computer vision and pattern recognition},
  pages={10965--10975},
  year={2022}
}

@article{gu2021open,
  title={Open-vocabulary object detection via vision and language knowledge distillation},
  author={Gu, Xiuye and Lin, Tsung-Yi and Kuo, Weicheng and Cui, Yin},
  journal={arXiv preprint arXiv:2104.13921},
  year={2021}
}

@inproceedings{radford2021learning,
  title={Learning transferable visual models from natural language supervision},
  author={Radford, Alec and Kim, Jong Wook and Hallacy, Chris and Ramesh, Aditya and Goh, Gabriel and Agarwal, Sandhini and Sastry, Girish and Askell, Amanda and Mishkin, Pamela and Clark, Jack and others},
  booktitle={International conference on machine learning},
  pages={8748--8763},
  year={2021},
  organization={PmLR}
}

@inproceedings{li2023blip,
  title={Blip-2: Bootstrapping language-image pre-training with frozen image encoders and large language models},
  author={Li, Junnan and Li, Dongxu and Savarese, Silvio and Hoi, Steven},
  booktitle={International conference on machine learning},
  pages={19730--19742},
  year={2023},
  organization={PMLR}
}

@article{vaswani2017attention,
  title={Attention is all you need},
  author={Vaswani, Ashish and Shazeer, Noam and Parmar, Niki and Uszkoreit, Jakob and Jones, Llion and Gomez, Aidan N and Kaiser, {\L}ukasz and Polosukhin, Illia},
  journal={Advances in neural information processing systems},
  volume={30},
  year={2017}
}

@article{dosovitskiy2020image,
  title={An image is worth 16x16 words: Transformers for image recognition at scale},
  author={Dosovitskiy, Alexey and Beyer, Lucas and Kolesnikov, Alexander and Weissenborn, Dirk and Zhai, Xiaohua and Unterthiner, Thomas and Dehghani, Mostafa and Minderer, Matthias and Heigold, Georg and Gelly, Sylvain and others},
  journal={arXiv preprint arXiv:2010.11929},
  year={2020}
}

@inproceedings{liu2021swin,
  title={Swin transformer: Hierarchical vision transformer using shifted windows},
  author={Liu, Ze and Lin, Yutong and Cao, Yue and Hu, Han and Wei, Yixuan and Zhang, Zheng and Lin, Stephen and Guo, Baining},
  booktitle={Proceedings of the IEEE/CVF international conference on computer vision},
  pages={10012--10022},
  year={2021}
}

@article{grattafiori2024llama,
  title={The llama 3 herd of models},
  author={Grattafiori, Aaron and Dubey, Abhimanyu and Jauhri, Abhinav and Pandey, Abhinav and Kadian, Abhishek and Al-Dahle, Ahmad and Letman, Aiesha and Mathur, Akhil and Schelten, Alan and Vaughan, Alex and others},
  journal={arXiv preprint arXiv:2407.21783},
  year={2024}
}

@article{team2024gemini,
  title={Gemini 1.5: Unlocking multimodal understanding across millions of tokens of context},
  author={Team, Gemini and Georgiev, Petko and Lei, Ving Ian and Burnell, Ryan and Bai, Libin and Gulati, Anmol and Tanzer, Garrett and Vincent, Damien and Pan, Zhufeng and Wang, Shibo and others},
  journal={arXiv preprint arXiv:2403.05530},
  year={2024}
}

@article{hoekman2017design,
  title={Design for ground beetle abundance and diversity sampling within the National Ecological Observatory Network},
  author={Hoekman, David and LeVan, Katherine E and Gibson, Cara and Ball, George E and Browne, Robert A and Davidson, Robert L and Erwin, Terry L and Knisley, C Barry and LaBonte, James R and Lundgren, Jonathan and others},
  journal={Ecosphere},
  volume={8},
  number={4},
  pages={e01744},
  year={2017},
  publisher={Wiley Online Library}
}

@article{blair2020robust,
  title={Robust and simplified machine learning identification of pitfall trap-collected ground beetles at the continental scale},
  author={Blair, Jarrett and Weiser, Michael D and Kaspari, Michael and Miller, Matthew and Siler, Cameron and Marshall, Katie E},
  journal={Ecology and evolution},
  volume={10},
  number={23},
  pages={13143--13153},
  year={2020},
  publisher={Wiley Online Library}
}

@article{wu2019deep,
  title={A deep learning model to recognize food contaminating beetle species based on elytra fragments},
  author={Wu, Leihong and Liu, Zhichao and Bera, Tanmay and Ding, Hongjian and Langley, Darryl A and Jenkins-Barnes, Amy and Furlanello, Cesare and Maggio, Valerio and Tong, Weida and Xu, Joshua},
  journal={Computers and Electronics in Agriculture},
  volume={166},
  pages={105002},
  year={2019},
  publisher={Elsevier}
}

@InProceedings{rayeed2025fine,
    author    = {Rayeed, S M and East, Alyson and Stevens, Samuel and Record, Synde and Stewart, Charles V.},
    title     = {Fine-Grained Beetle Taxonomy with Vision Models: A Benchmark on Long-Tailed and Domain-Adaptive Classification},
    booktitle = {Proceedings of the IEEE/CVF International Conference on Computer Vision (ICCV) Workshops},
    month     = {October},
    year      = {2025},
    pages     = {5093-5099}
}

@article{rayeed2025beetleverse,
    title = {BeetleVerse: A study on taxonomic classification of ground beetles},
    author = {Rayeed, SM and East, Alyson and Stevens, Samuel and Record, Synde and Stewart, Charles V.},
    journal={IEEE/CVF Conference on Computer Vision and Pattern Recognition (CVPR) Workshops},
    year={2025},
    eprint={2504.13393},
    archivePrefix={arXiv},
    primaryClass={cs.CV},
    url={https://arxiv.org/abs/2504.13393}, 
}

@article{east2025optimizing,
  title={Optimizing image capture for computer vision-powered taxonomic identification and trait recognition of biodiversity specimens},
  author={East, Alyson and Campolongo, Elizabeth G and Meyers, Luke and Rayeed, SM and Stevens, Samuel and Zarubiieva, Iuliia and Fluck, Isadora E and Gir{\'o}n, Jennifer C and Jousse, Maximiliane and Lowe, Scott and others},
  journal={Methods in Ecology and Evolution},
  year={2025},
  publisher={Wiley Online Library}
}

@misc{NEON-pinned-beetles,
  doi = {10.48443/CD21-Q875},
  url = {https://data.neonscience.org/data-products/DP1.10022.001/RELEASE-2025},
  author = {{National Ecological Observatory Network (NEON)}},
  language = {en},
  title = {Ground beetles sampled from pitfall traps (DP1.10022.001)},
  publisher = {National Ecological Observatory Network (NEON)},
  year = {2025}
}

@misc{NEON-field-metadata,
  url = {https://www.neonscience.org/field-sites/exports/NEON_Field_Site_Metadata_20250625},
  author = {{{National Ecological Observatory Network (NEON)}}},
  language = {en},
  title = {{NEON} Field Site Metadata},
  publisher = {National Ecological Observatory Network (NEON)},
  year = {2025},
  note = {Dataset accessed from https://data.neonscience.org/api/v0/locations/sites on June 25, 2025}
}

@misc{Fluck2018_NEON_Beetle,
  author = {Isadora E. Fluck and Benjamin Baiser and Riley Wolcheski and Isha Chinniah and Sydne Record},
  title = {2018 {NEON} Ethanol-preserved Ground Beetles (Revision 7b3731d)},
  year = {2025},
  url = {https://huggingface.co/datasets/imageomics/2018-NEON-beetles},
  doi = {10.57967/hf/5252},
  publisher = {Hugging Face}
}

@software{Ramirez_2018_NEON_Ethanol-preserved_2025,
        author = {Ramirez, Michelle and Nepovinnykh, Ekaterina and Ali, Sarwan and Campolongo, Elizabeth G.},
        doi = {10.5281/zenodo.16989738},
        license = {MIT},
        title = {{2018 NEON Ethanol-preserved Ground Beetles Processing}},
        url = {https://github.com/Imageomics/2018-NEON-beetles-processing},
        version = {2.0.0},
        year = {2025}
}
\bibliographystyle{IEEEtran}

%%%%%%%%%%%%%%%%%%%%%%%%%%%%%%%%%%%%%%%%%%%%%%%%%%%%%%%%%%%%

% \newpage
% \appendix
% \input{appendix}

\end{document}